\documentclass[11pt,a4paper]{article}
\usepackage[hyperref]{acl2018}
\usepackage{times}
\usepackage{amsmath}
\usepackage{latexsym}
\usepackage{graphicx}
\usepackage{url}
\usepackage{amssymb}
\usepackage{color}
\usepackage{multirow}
\usepackage{makecell}

\DeclareMathOperator*{\argmax}{\text{arg\,max}}

\newcommand{\Cc}{\mathcal{C}}
\newcommand{\Dc}{\mathcal{D}}

\newcommand{\Sc}{\mathcal{S}}

\newcommand{\Yc}{\mathcal{Y}}

\newcommand{\vv}{\mathbf{v}}

\newcommand{\xv}{\mathbf{x}}
\newcommand{\yv}{\mathbf{y}}

\aclfinalcopy 
\def\aclpaperid{1502} 

\title{Neural Factor Graph Models for Cross-lingual Morphological Tagging}

\author{Chaitanya Malaviya \and Matthew R. Gormley \and Graham Neubig 
  \\ {Language Technologies Institute, Machine Learning Department} \\ {Carnegie Mellon University} \\
  {\tt \{cmalaviy,mgormley,gneubig\}@cs.cmu.edu}}

\date{}

\begin{document}
\maketitle
\begin{abstract}
 Morphological analysis involves predicting the syntactic traits of a word (e.g. \{\textit{POS: Noun, Case: Acc, Gender: Fem}\}). Previous work in morphological tagging improves performance for low-resource languages (LRLs) through cross-lingual training with a high-resource language (HRL) from the same family, but is limited by the strict---often false---assumption that tag sets exactly overlap between the HRL and LRL. In this paper we propose a method for cross-lingual morphological tagging that aims to improve information sharing between languages by relaxing this assumption. The proposed model uses factorial conditional random fields with neural network potentials, making it possible to (1) utilize the expressive power of neural network representations to smooth over superficial differences in the surface forms, (2) model pairwise and transitive relationships between tags, and (3) accurately generate tag sets that are unseen or rare in the training data. Experiments on four languages from the Universal Dependencies Treebank~\cite{ud2.1} demonstrate superior tagging accuracies over existing cross-lingual approaches.\footnote{Our code and data is publicly available at \texttt{www.github.com/chaitanyamalaviya/\\NeuralFactorGraph}.}
\end{abstract}

\section{Introduction}

\begin{figure}[h]
  \centering
  \includegraphics[width=7cm]{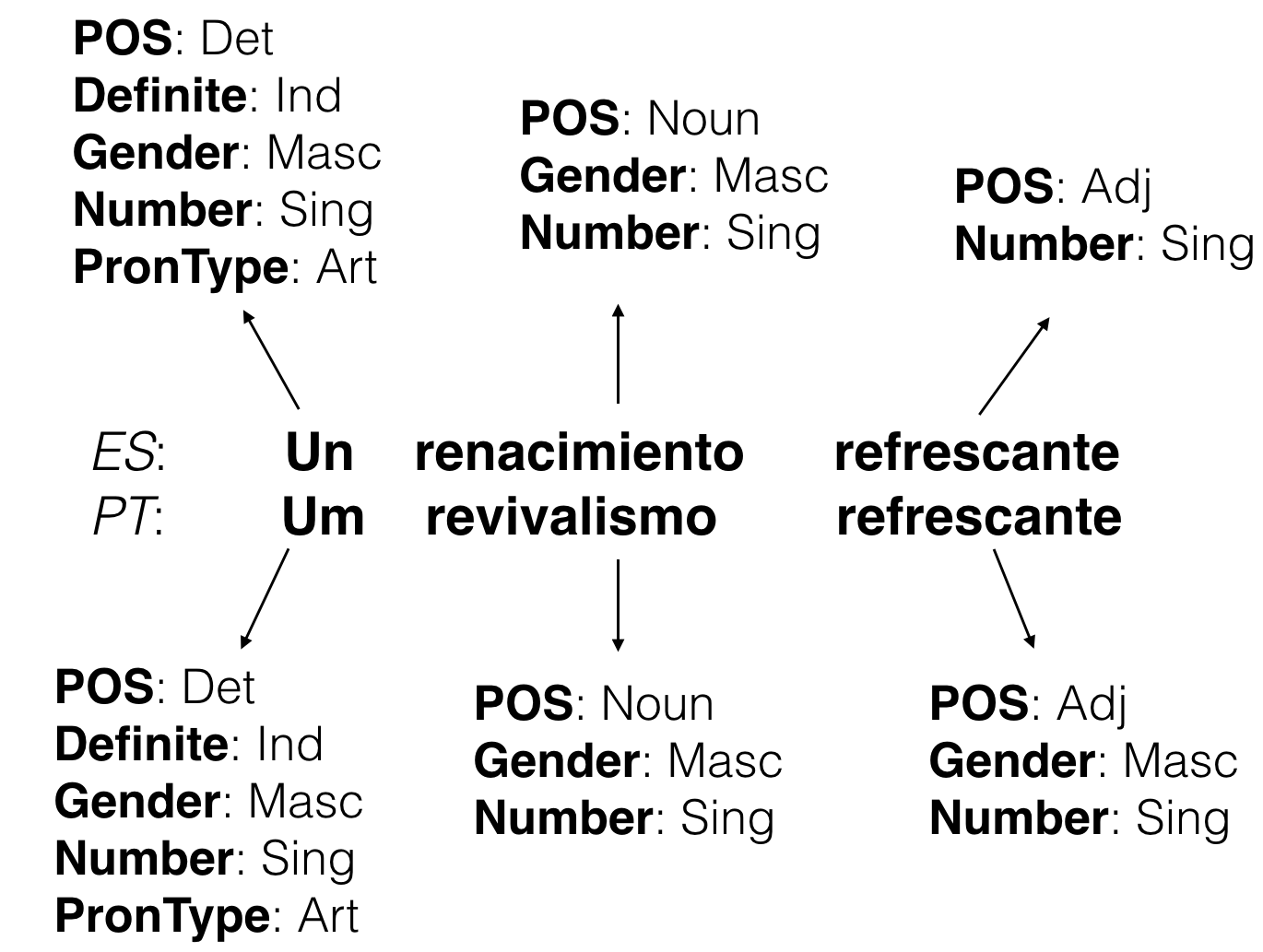}
  \caption{\label{fig:tags} Morphological tags for a UD sentence in Portuguese and a translation in Spanish}
\end{figure}

Morphological analysis (\newcite{hajivc1998tagging}, \newcite{oflazer1994tagging}, \emph{inter alia}) is the task of predicting fine-grained annotations about the syntactic properties of tokens in a language such as part-of-speech, case, or tense.
For instance, in Figure~\ref{fig:tags}, the given Portuguese sentence is labeled with the respective morphological tags such as Gender and its label value Masculine.


The accuracy of morphological analyzers is paramount, because their results are often a first step in the NLP pipeline for tasks such as translation~\cite{vylomova2017,tsarfaty2010statistical} and parsing \cite{tsarfaty2013parsing}, and errors in the upstream analysis may cascade to the downstream tasks.
One difficulty, however, in creating these taggers is that only a limited amount of annotated data is available for a majority of the world's languages to learn these morphological taggers.
Fortunately, recent efforts in morphological annotation follow a standard annotation schema for these morphological tags across languages, and now the Universal Dependencies Treebank~\cite{ud2.1} has tags according to this schema in 60 languages.

\newcite{cotterell2017crossling} have recently shown that combining this shared schema with cross-lingual training on a related high-resource language (HRL) gives improved performance on tagging accuracy for low-resource languages (LRLs). The output space of this model consists of tag sets such as \{\textit{POS: Adj, Gender: Masc, Number: Sing}\}, which are predicted for a token at each time step.
However, this model relies heavily on the fact that \emph{the entire space of tag sets} for the LRL must match those of the HRL, which is often not the case, either due to linguistic divergence or small differences in the annotation schemes between the two languages.%
\footnote{In particular, the latter is common because many UD resources were created by full or semi-automatic conversion from treebanks with less comprehensive annotation schemes than UD. Our model can generate label values for these tags too, which could possibly aid the enhancement of UD annotations, although we do not examine this directly in our work.}
For instance, in Figure~\ref{fig:tags} ``refrescante'' is assigned a gender in the Portuguese UD treebank, but not in the Spanish UD treebank.


\begin{figure}[t!]
  \centering
  \includegraphics[width=7.5cm]{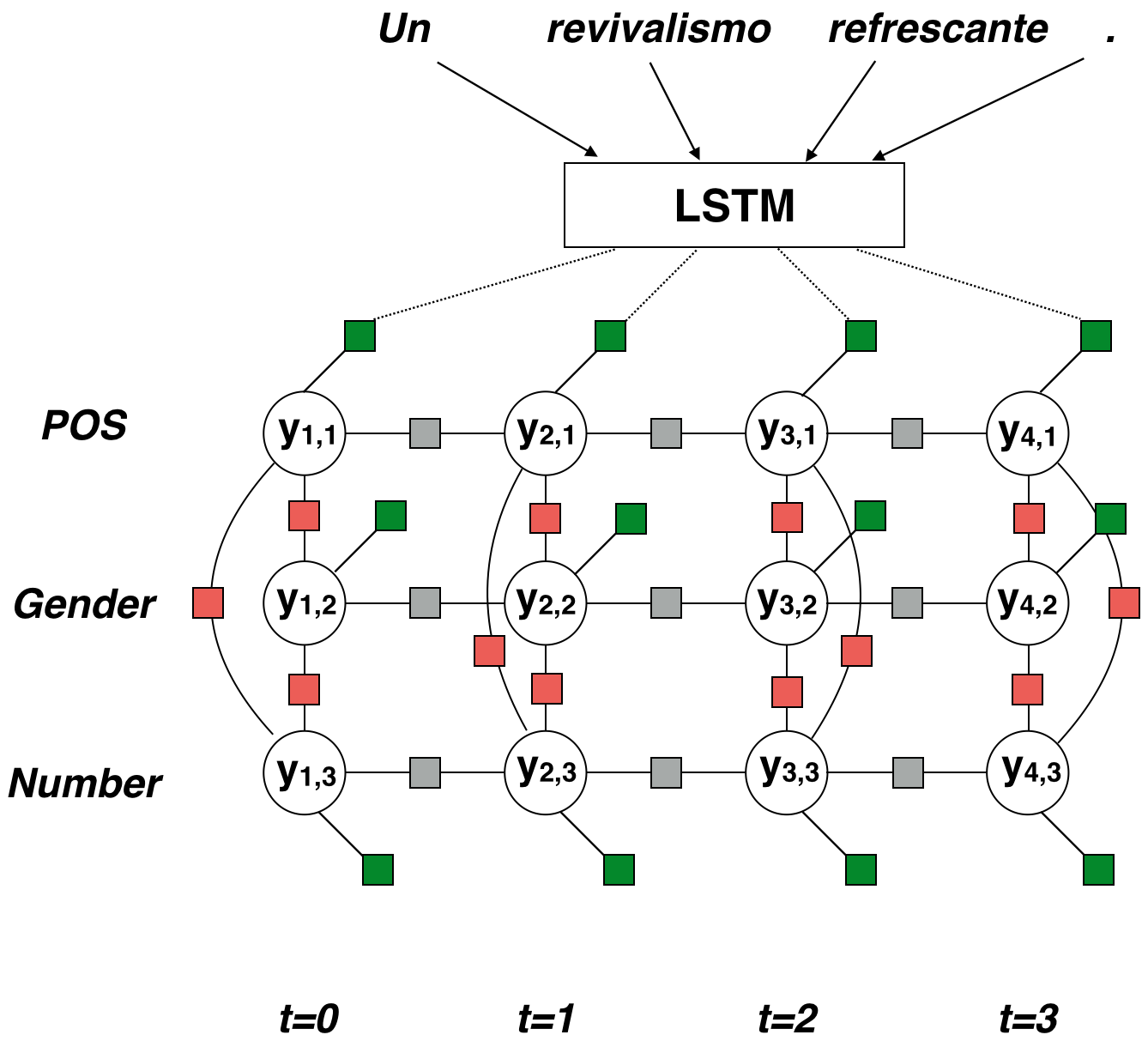}
  \caption{\label{fig:model} FCRF-LSTM Model for morphological tagging }
\end{figure}

In this paper, we propose a method that instead of predicting full tag sets, makes predictions over single tags separately but ties together each decision by modeling variable dependencies between tags over time steps (e.g. capturing the fact that nouns frequently occur after determiners) and pairwise dependencies between all tags at a single time step (e.g. capturing the fact that infinitive verb forms don't have tense).
The specific model is shown in Figure~\ref{fig:model}, consisting of a factorial conditional random field (FCRF; \newcite{sutton2007dynamic}) with neural network potentials calculated by long short-term memory (LSTM; \cite{hochreiter1997long}) at every variable node (\S\ref{sec:model}).
Learning and inference in the model is made tractable through belief propagation over the possible tag combinations, allowing the model to consider an exponential label space in polynomial time (\S\ref{sec:inference}).

This model has several advantages:
\begin{itemize}
\item{The model is able to generate tag sets unseen in training data, and share information between similar tag sets, alleviating the main disadvantage of previous work cited above.}
\item{Our model is empirically strong, as validated in our main experimental results: it consistently outperforms previous work in cross-lingual low-resource scenarios in experiments.}
\item{Our model is more interpretable, as we can probe the model parameters to understand which variable dependencies are more likely to occur in a language, as we demonstrate in our analysis.}
\end{itemize}
In the following sections, we describe the model and these results in more detail.





\section{Problem Formulation and Baselines}
\label{sec:problem}

\subsection{Problem Formulation}

Formally, we define the problem of morphological analysis as the task of mapping a length-$T$ string of tokens $\xv = x_1, \ldots, x_T$ into the target morphological tag sets for each token $\yv = \yv_1, \ldots, \yv_T$.
For the $t$th token, the target label $\yv_t = y_{t,1}, \ldots, y_{t,m}$ defines a set of tags (e.g. \{\textit{Gender: Masc, Number: Sing, POS: Verb}\}).
An annotation schema defines a set $\Sc$ of $M$ possible tag types and with the $m$th type (e.g. {\it Gender}) defining its set of possible labels $\Yc_m$ (e.g. {\it \{Masc, Fem, Neu\}}) such that $y_{t,m} \in \Yc_m$.
We must note that not all tags or attributes need to be specified for a token; usually, a subset of $\Sc$ is specified for a token and the remaining tags can be treated as mapping to a $\texttt{NULL} \in \Yc_m$ value.
Let $\Yc = \{(y_{1}, \ldots, y_{M}) : y_1 \in \Yc_1, \ldots, y_M \in \Yc_M \}$ denote the set of all possible tag sets.

\subsection{Baseline: Tag Set Prediction}


Data-driven models for morphological analysis are constructed using training data $\Dc = \{ (\xv^{(i)}, \yv^{(i)}) \}_{i=1}^N$ consisting of $N$ training examples. 
The baseline model~\cite{cotterell2017crossling} we compare with  regards the output space of the model as a subset $\tilde{\Yc} \subset \Yc$ where $\tilde{\Yc}$ is the set of all tag sets seen in this training data. Specifically, they solve the task as a multi-class classification problem where the classes are individual tag sets. In low-resource scenarios, this indicates that $|\tilde{\Yc}| << |\Yc|$ and even for those tag sets existing in $\tilde{\Yc}$ we may have seen very few training examples. The conditional probability of a sequence of tag sets given the sentence is formulated as a $0$th order CRF.
\begin{equation}
p(\yv|\xv)=\prod_{t=1}^T p(\yv_t | \xv)
\label{eq:baseline}
\end{equation}

Instead, we would like to be able to generate any combination of tags from the set $\Yc$, and share statistical strength among similar tag sets. 

\subsection{A Relaxation: Tag-wise Prediction}
\label{sec:tagwise}


As an alternative, we could consider a model that performs prediction for each tag's label $y_{t,m}$ independently.
\begin{equation}
p(\yv|\xv)=\prod_{t=1}^T \prod_{m=1}^{M} p(y_{t,m} | \xv)
\label{eq:tagwise}
\end{equation}
This formulation has an advantage: the tag-predictions within a single time step are now independent, it is now easy to generate any combination of tags from $\Yc$.
On the other hand, now it is difficult to model the interdependencies between tags in the same tag set $\yv_i$, a major disadvantage over the previous model.
In the next section, we describe our proposed neural factor graph model, which can model not only dependencies within tags for a single token, but also dependencies across time steps while still maintaining the flexibility to generate any combination of tags from $\Yc$.



\section{Neural Factor Graph Model}
\label{sec:model}

Due to the correlations between the syntactic properties that are represented by morphological tags, we can imagine that capturing the relationships between these tags through pairwise dependencies can inform the predictions of our model.
These dependencies exist both among tags for the same token (intra-token pairwise dependencies), and across tokens in the sentence (inter-token transition dependencies).
For instance, knowing that a token's POS tag is a Noun, would strongly suggest that this token would have a $\texttt{NULL}$ label for the tag Tense, with very few exceptions \cite{nordlinger04nominaltense}.
In a language where nouns follow adjectives, a tag set prediction \textit{\{POS: Adj, Gender: Fem\}} might inform the model that the next token is likely to be a noun and have the same gender. The baseline model can not explicitly model such interactions given their factorization in equation~\ref{eq:baseline}.

To incorporate the dependencies discussed above, we define a factorial CRF~\cite{sutton2007dynamic}, with pairwise links between cotemporal variables and transition links between the same types of tags. This model defines a distribution over the tag-set sequence $\yv$ given the input sentence $\xv$ as,
\begin{equation}
p(\yv|\xv)=\frac{1}{Z(\xv)} \prod_{t=1}^{T} \prod_{\alpha \in \Cc} \psi_{\alpha}(\yv_{\alpha}, \xv,t)
\end{equation}
where $\Cc$ is the set of factors in the factor graph (as shown in Figure \ref{fig:model}), $\alpha$ is one such factor, and $\yv_{\alpha}$ is the assignment to the subset of variables neighboring factor $\alpha$.
We define three types of potential functions: neural $\psi_{NN}$, pairwise $\psi_{P}$, and transition $\psi_{T}$, described in detail below. 

\begin{figure}[h]
  \centering
  \includegraphics[width=3.8cm]{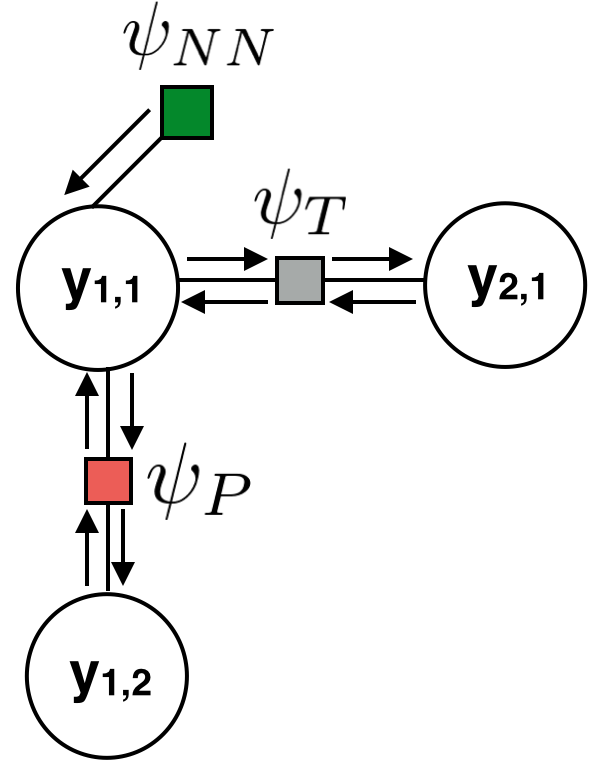}
  \caption{\label{fig:graph} Factors in the Neural Factor Graph model (red: Pairwise, grey: Transition, green: Neural Network) }
\end{figure}

\subsection{Neural Factors}

The flexibility of our formulation allows us to include any form of custom-designed potentials in our model. Those for the neural factors have a fairly standard log-linear form, 
\begin{equation}
\psi_{{NN},i}(y_{t,m}) =\exp{ \left\{ \sum_{k} \lambda_{\textnormal{nn},k} f_{\textnormal{nn},k}(\xv, t) \right\} }
\end{equation}
except that the features $f_{\textnormal{nn},k}$ are themselves given by a neural network. There is one such factor per variable.
We obtain our neural factors using a biLSTM over the input sequence $\xv$, where the input word embedding for each token is obtained from a character-level biLSTM embedder. This component of our model is similar to the model proposed by \citet{cotterell2017crossling}. Given an input token  $\xv_t=c_1...c_n$, we compute an input embedding $v_t$ as,
\begin{equation}
v_t=[\textnormal{cLSTM}(c_1...c_n) ; \textnormal{cLSTM}(c_n...c_1)]
\label{eq:charLSTM}
\end{equation}
Here, $\textnormal{cLSTM}$ is a character-level LSTM function that returns the last hidden state. This input embedding $v_t$ is then used in the biLSTM tagger to compute an output representation $e_t$. Finally, the scores $f_{\textnormal{nn}}(\xv,t)$ are obtained as,
\begin{equation}
f_{\textnormal{nn}}(\xv,t)=W_l e_t + b_l
\label{eq:nnscores}
\end{equation}
We use a language-specific linear layer with weights $W_l$ and bias $b_l$.

\subsection{Pairwise Factors}
As discussed previously, the pairwise factors are crucial for modeling correlations between tags. The pairwise factor potential for a tag $i$ and tag $j$ at timestep $t$ is given in equation~\ref{eq:pair}. Here, the dimension of $f_{\textnormal{p}}$ is $(|\Yc_i|, |\Yc_j|)$.
These scores are used to define the neural factors as,
\begin{equation}
\psi_{P_{i,j}}(y_{t,i}, y_{t,j}) =\exp{ \left\{ \sum_{k} \lambda_{\textnormal{p},k} f_{\textnormal{p},k}(y_{t,i}, y_{t,j}) \right\} }
\label{eq:pair}
\end{equation}

\subsection{Transition Factors}
Previous work has experimented with the use of a linear chain CRF with factors from a neural network \cite{huang2015bidirectional} for sequence tagging tasks. We hypothesize that modeling transition factors in a similar manner can allow the model to utilize information about neighboring tags and capture word order features of the language. The transition factor for tag $i$ and timestep $t$ is given below for variables $y_{t,i}$ and $y_{t+1,i}$. The dimension of $f_{\textnormal{T}}$ is $(|\Yc_i|, |\Yc_i|)$.
\begin{equation}
\small
\psi_{T_{i,t}}(y_{t,i}, y_{t+1,i}) =\exp{ \left\{ \sum_{k} \lambda_{\textnormal{T},k} f_{\textnormal{T},k}(y_{t,i}, y_{t+1,i}) \right\} }
\end{equation}
In our experiments, $f_{\textnormal{p},k}$ and $f_{\textnormal{T},k}$ are simple indicator features for the values of tag variables with no dependence on $\xv$.

\subsection{Language-Specific Weights}
As an enhancement to the information encoded in the transition and pairwise factors, we experiment with training general and language-specific parameters for the transition and the pairwise weights. We define the weight matrix $\lambda_{\textnormal{gen}}$ to learn the general trends that hold across both languages, and the weights $\lambda_{\textnormal{lang}}$ to learn the exceptions to these trends. In our model, we sum both these parameter matrices before calculating the transition and pairwise factors. For instance, the transition weights $\lambda_T$ are calculated as $\lambda_T=\lambda_{\textnormal{T, gen}} + \lambda_{\textnormal{T, lang}}$.

\subsection{Loopy Belief Propagation}
\label{sec:inference}

Since the graph from Figure~\ref{fig:model} is a loopy graph, performing exact inference can be expensive. Hence, we use loopy belief propagation \cite{murphy1999loopy,ihler2005loopy} for computation of approximate variable and factor marginals. Loopy BP is an iterative message passing algorithm that sends messages between variables and factors in a factor graph. The message updates from variable $v_i$, with neighboring factors $N(i)$, to factor $\alpha$ is
%
\begin{equation}
 \mu_{i \rightarrow \alpha} (v_i) = \prod_{\alpha \in N(i) \backslash \alpha} \mu_{\alpha \rightarrow i} (v_i)
\end{equation}
The message from factor $\alpha$ to variable $v_i$ is
\begin{equation}
 \small
 \mu_{\alpha \rightarrow i} (v_i) = \sum_{\vv_\alpha : \vv_\alpha [i]=v_i} \psi_\alpha (\vv_\alpha) \prod_{j \in N(\alpha) \backslash i} \mu_{j \rightarrow \alpha} (\vv_\alpha [i])
\end{equation}
where $\vv_{\alpha}$ denote an assignment to the subset of variables adjacent to factor $\alpha$, and $\vv_{\alpha}[i]$ is the assignment for variable $v_i$. 
Message updates are performed asynchronously in our model.
Our message passing schedule was similar to that of foward-backward: the forward pass sends all messages from the first time step in the direction of the last. Messages to/from pairwise factors are included in this forward pass. The backward pass sends messages in the direction from the last time step back to the first. This process is repeated until convergence.
%
%
%
We say that BP has converged when the maximum residual error \cite{sutton07rbp0} over all messages is below some threshold.
%
%
Upon convergence, we obtain the belief values of variables and factors as,
\begin{align}
 b_{i} (v_i) &= \frac{1}{\kappa_i} \prod_{\alpha \in N(i)} \mu_{\alpha \rightarrow i} (v_i) \\
 b_{\alpha} (v_\alpha) &= \frac{1}{\kappa_{\alpha}} \psi_\alpha(v_\alpha) \prod_{i \in N(\alpha)} \mu_{i \rightarrow \alpha} (v_\alpha [i])
\end{align}
where $\kappa_i$ and $\kappa_{\alpha}$ are normalization constants ensuring that the beliefs for a variable $i$ and factor $\alpha$ sum-to-one. In this way, we can use the beliefs as approximate marginal probabilities. 

\subsection{Learning and Decoding}
\label{sec:learning}

We perform end-to-end training of the neural factor graph by following
the (approximate) gradient of the log-likelihood
$\sum_{i=1}^N \log p(\yv^{(i)} | \xv^{(i)})$. The true gradient
requires access to the marginal probabilities for each factor,
e.g. $p(\yv_{\alpha} | \xv)$ where $\yv_{\alpha}$ denotes the subset
of variables in factor $\alpha$. For example, if $\alpha$ is a transition factor for tag $m$ at timestep $t$, then $\yv_{\alpha}$ would be $y_{t,m}$ and $y_{t+1,m}$. Following \cite{sutton2007dynamic},
we replace these marginals with the beliefs $b_{\alpha}(\yv_{\alpha})$
from loopy belief propagation.\footnote{Using this approximate gradient is
  akin to the \emph{surrogate likelihood} training of
  \cite{wainwright2006estimating}.}
Consider the log-likelihood of a single example $\ell^{(i)} = \log p(\yv^{(i)} | \xv^{(i)})$.
The partial derivative with respect to parameter $\lambda_{g,k}$ for
each type of factor $g \in \{NN, T, P\}$ is the difference of the observed features with the expected features under the model's (approximate) distribution as represented by the beliefs:
%
%
\begin{align*}
  \resizebox{\columnwidth}{!}{
  $\displaystyle
  \frac{\partial \ell^{(i)}}{\partial \lambda_{g,k} } = 
  \sum_{\alpha \in \Cc_g} \left( f_{g,k}(\yv_{\alpha}^{(i)}) - \sum_{\yv_{\alpha}} b_{\alpha}(\yv_{\alpha}) f_{g,k}(\yv_\alpha) \right)
  $
  }
\end{align*}
where $\Cc_g$ denotes all the factors of type $g$, and we have omitted
any dependence on $\xv^{(i)}$ and $t$ for brevity---$t$ is accessible through the factor index $\alpha$. 
For the neural network factors, the features are given by a
biLSTM. We backpropagate through to the biLSTM parameters using the
partial derivative below,
\begin{align*}
  \frac{\partial \ell^{(i)}}{\partial f_{NN,k}(y_{t,m}^{(i)}, t) } = \lambda_{NN,k} - \sum_{y_{t,m}} b_{t,m}(y_{t,m}) \lambda_{NN,k}
\end{align*}
where $b_{t,m}(\cdot)$ is the variable belief corresponding to
variable $y_{t,m}$.

%


To predict a sequence of tag sets $\hat{\yv}$ at test time, we use
minimum Bayes risk (MBR) decoding \cite{bickel1977mathematical,goodman1996efficient} for Hamming loss over
tags. For a variable $y_{t,m}$ representing tag $m$ at timestep $t$,
we take
\begin{equation}
\hat{y}_{t,m} = \argmax_{l \in \Yc_m} b_{t,m}(l).
\end{equation}
where $l$ ranges over the possible labels for tag $m$.

\section{Experimental Setup}
\label{sec:experiments}

\begin{table}[t]
\begin{center}
\begin{tabular}{|c|c|c|c|}
\hline \bf Language Pair & \bf  HRL Train & \bf Dev & \bf Test  \\ \hline 
\textsc{da/sv} & 4,383 & 504 & 1219 \\ \hline
\textsc{ru/bg} & 3,850 & 1115 & 1116 \\ \hline
\textsc{fi/hu} & 12,217 & 441 & 449 \\ \hline
\textsc{es/pt} & 14,187 & 560 & 477 \\ \hline
\end{tabular}
\end{center}
\caption{\label{dataset_size} Dataset sizes. $tgt\_size=100$ or 1,000 LRL sentences are added to HRL Train}
\end{table}

\begin{table}[t]
\begin{center}
\begin{tabular}{|c|c|c|}
\hline \bf Language Pair & \bf  Unique Tags & \bf Tag Sets  \\ \hline 
\textsc{da/sv} & 23 & 224 \\ \hline
\textsc{ru/bg} & 19 & 798 \\ \hline
\textsc{fi/hu} & 27 & 2195 \\ \hline
\textsc{es/pt} & 19  & 451 \\ \hline
\end{tabular}
\end{center}
\caption{\label{tagset_size} Tag Set Sizes with $tgt\_size$=100}
\end{table} 

\begin{table*}[t]
\begin{center}
\begin{tabular}{|c|c|c@{\hskip 0.1in}c@{\hskip 0.1in}c | c@{\hskip 0.1in}c@{\hskip 0.1in}c|}
\hline
\multirow{2}{*}{Language} &  \multirow{2}{*}{Model} & \multicolumn{3}{c|}{$tgt\_size$ = 100} & \multicolumn{3}{c|}{$tgt\_size$=1000}  \\
   & & Accuracy & F1-Macro & F1-Micro & Accuracy & F1-Macro & F1-Micro \\
\hline \hline
 \multirow{2}{*}{\textsc{sv}} & Baseline  & 15.11 & 8.36 & 10.37 & 68.64 & 76.36 & 76.50 \\
  & Ours & \textbf{29.47} &\textbf{ 54.09} & \textbf{54.36} & \textbf{71.32} & \textbf{84.42} & \textbf{84.46} \\
\hline
 \multirow{2}{*}{\textsc{bg}}  & Baseline & \textbf{29.05} & 14.32 & 29.62 & \textbf{59.20} & \textbf{67.22} & \textbf{67.12} \\
  & Ours & 27.81 & \textbf{40.97} & \textbf{42.43 }& 39.25 & 60.23 & 60.84 \\
\hline
 \multirow{2}{*}{\textsc{hu}}  & Baseline  & 21.97 & 13.30 & 16.67 & \textbf{50.75} & 58.68 & 62.79 \\
  & Ours & \textbf{33.32} & \textbf{54.88} & \textbf{54.69} & 45.90 & \textbf{74.05 }& \textbf{73.38 }\\
\hline
 \multirow{2}{*}{\textsc{pt}} & Baseline & 18.91 & 7.10 & 10.33 & 74.22 & 81.62 & 81.87 \\
  & Ours & \textbf{58.82} & \textbf{73.67} & \textbf{74.07} & \textbf{76.26} & \textbf{87.13} & \textbf{87.22}  \\
\hline
\end{tabular}
\end{center}
\caption{\label{tab:mono} Token-wise accuracy and F1 scores on mono-lingual experiments}
\end{table*} 

\subsection{Dataset}

We used the Universal Dependencies Treebank UD v2.1 \cite{ud2.1} for our experiments. 
We picked four low-resource/high-resource language pairs, each from a different family: Danish/Swedish (\textsc{da/sv}), Russian/Bulgarian  (\textsc{ru/bg}), Finnish/Hungarian (\textsc{fi/hu}), Spanish/Portuguese (\textsc{es/pt}). Picking languages from different families would ensure that we obtain results that are on average consistent across languages.

The sizes of the training and evaluation sets are specified in Table~\ref{dataset_size}.
In order to simulate low-resource settings, we follow the experimental procedure from~\citet{cotterell2017crossling}. We restrict the number of sentences of the target language ($tgt\_size$) in the training set to 100 or 1000 sentences. We also augment the tag sets in our training data by adding a $\texttt{NULL}$ label for all tags that are not seen for a token. It is expected that our model will learn which tags are unlikely to occur given the variable dependencies in the factor graph.
The dev set and test set are only in the target language. From Table~\ref{tagset_size}, we can see there is also considerable variance in the number of unique tags and tag sets found in each of these language pairs.

\subsection{Baseline Tagger}

As the baseline tagger model, we re-implement the \textsc{specific} model from ~\citet{cotterell2017crossling} that uses a language-specific softmax layer. Their model architecture uses a character biLSTM embedder to obtain a vector representation for each token, which is used as input in a word-level biLSTM. The output space of their model is all the tag sets seen in the training data. This work achieves strong performance on several languages from UD on the task of morphological tagging and is a strong baseline.

\subsection{Training Regimen}

We followed the parameter settings from ~\citet{cotterell2017crossling} for the baseline tagger and the neural component of the \textsc{FCRF-LSTM} model. For both models, we set the input embedding and linear layer dimension to 128. We used 2 hidden layers for the LSTM where the hidden layer dimension was set to 256 and a dropout ~\cite{srivastava2014dropout} of 0.2 was enforced during training.
All our models were implemented in the PyTorch toolkit \cite{paszke2017automatic}. The parameters of the character biLSTM and the word biLSTM were initialized randomly. We trained the baseline models and the neural factor graph model with SGD and Adam respectively for 10 epochs each, in batches of 64 sentences. These optimizers gave the best performances for the respective models.

For the FCRF, we initialized transition and pairwise parameters with zero weights, which was important to ensure stable training. We considered BP to have reached convergence when the maximum residual error was below 0.05 or if the maximum number of iterations was reached (set to 40 in our experiments). 
We found that in cross-lingual experiments, when $tgt\_size=100$, the relatively large amount of data in the HRL was causing our model to overfit on the HRL and not generalize well to the LRL. As a solution to this, we upsampled the LRL data by a factor of 10 when $tgt\_size=100$ for both the baseline and the proposed model.

\begin{table*}[t]
\begin{center}
\begin{tabular}{|c|c|c@{\hskip 0.1in}c@{\hskip 0.1in}c | c@{\hskip 0.1in}c@{\hskip 0.1in}c|}
\hline
\multirow{2}{*}{Language} &  \multirow{2}{*}{Model} & \multicolumn{3}{c|}{$tgt\_size$ = 100} & \multicolumn{3}{c|}{$tgt\_size$=1000}  \\
   & & Accuracy & F1-Macro & F1-Micro & Accuracy & F1-Macro & F1-Micro \\
\hline \hline
 \multirow{2}{*}{\textsc{da/sv}} & Baseline  & \textbf{66.06} & 73.95 & 74.37 & \textbf{82.26} &\textbf{ 87.88} &\textbf{ 87.91} \\
  & Ours & 63.22 & \textbf{78.75} & \textbf{78.72} & 77.43 & 87.56 & 87.52 \\
\hline
 \multirow{2}{*}{\textsc{ru/bg}}  & Baseline & \textbf{52.76 }& 58.41 & 58.23 & \textbf{71.90} & 77.89 & 77.97 \\
  & Ours & 46.89 & \textbf{64.46 }& \textbf{64.75 }& 67.56 & \textbf{82.06} & \textbf{82.11} \\
\hline
 \multirow{2}{*}{\textsc{fi/hu}}  & Baseline & \textbf{51.74} & 68.15 & 66.82 & 61.80 & 75.96 & 76.16\\
  & Ours & 45.41 & \textbf{68.63} & \textbf{68.07} & \textbf{63.93} & \textbf{85.06} & \textbf{84.12} \\
\hline
 \multirow{2}{*}{\textsc{es/pt}} & Baseline & \textbf{79.40} & 86.03 & 86.14 & \textbf{85.85 }& 91.91 & 91.93 \\
  & Ours & 77.75 & \textbf{88.42} & \textbf{88.44} & 85.02 & \textbf{92.35} & \textbf{92.37} \\
\hline
\end{tabular}
\end{center}
\caption{\label{tab:cross} Token-wise accuracy and F1 scores on cross-lingual experiments }
\end{table*}

\paragraph{Evaluation:} Previous work on morphological analysis \cite{cotterell2017crossling,buys-botha:2016:P16-1} has reported scores on average token-level accuracy and F1 measure. The average token level accuracy counts a tag set prediction as correct only it is an exact match with the gold tag set. On the other hand, F1 measure is measured on a tag-by-tag basis, which allows it to give partial credit to partially correct tag sets. Based on the characteristics of each evaluation measure, Accuracy will favor tag-set prediction models (like the baseline), and F1 measure will favor tag-wise prediction models (like our proposed method). Given the nature of the task, it seems reasonable to prefer getting some of the tags correct (e.g. Noun+Masc+Sing becomes Noun+Fem+Sing), instead of missing all of them (e.g. Noun+Masc+Sing becomes Adj+Fem+Plur). F-score gives partial credit for getting some of the tags correct, while tagset-level accuracy will treat these two mistakes equally. Based on this, we believe that F-score is intuitively a better metric. However, we report both scores for completeness.
\label{para:evaluation}

\section{Results and Analysis}
\label{sec:analysis}




\subsection{Main Results}


First, we report the results in the case of monolingual training in Table \ref{tab:mono}. The first row for each language pair reports the results for our reimplementation of \newcite{cotterell2017crossling}, and the second for our full model.
From these results, we can see that we obtain improvements on the F-measure over the baseline method in most experimental settings except \textsc{BG} with $tgt\_size=1000$.
In a few more cases, the baseline model sometimes obtains higher accuracy scores for the reason described in~\ref{para:evaluation}.

In our cross-lingual experiments shown in Table \ref{tab:cross}, we also note F-measure improvements over the baseline model with the exception of \textsc{DA/SV} when $tgt\_size=1000$. We observe that the improvements are on average stronger when $tgt\_size=100$. This suggests that our model performs well with very little data due to its flexibility to generate any tag set, including those not observed in the training data. The strongest improvements are observed for \textsc{FI/HU}. This is likely because the number of unique tags is the highest in this language pair and our method scales well with the number of tags due to its ability to make use of correlations between the tags in different tag sets.

\begin{table}[h]
\begin{center}
\begin{tabular}{|c|c@{\hskip 0.1in}c|c|}
\hline Language & Transition & Pairwise & F1-Macro\\ \hline
\multirow{4}{*}{\textsc{hu}}& $\times$ & $\times$  & 69.87 \\
 & $\checkmark$ & $\times$  & 73.21 \\
 & $\times$ & $\checkmark$ & 73.68\\
 & $\checkmark$ & $\checkmark$  & 74.05 \\ \hline
\multirow{3}{*}{\textsc{fi/hu}} & $\times$ & $\times$  & 79.57 \\
 & $\checkmark$ & $\times$ & 84.41 \\
 & $\times$ & $\checkmark$ & 84.73 \\
 & $\checkmark$ & $\checkmark$ & 85.06 \\ \hline
\end{tabular}
\end{center}
\caption{\label{tab:ablation} Ablation Experiments ($tgt\_size$=1000)} 
\end{table} 

To examine the utility of our transition and pairwise factors, we also report results on ablation experiments by removing transition and pairwise factors completely from the model in Table~\ref{tab:ablation}.
Ablation experiments for each factor showed decreases in scores relative to the model where both factors are present, but the decrease attributed to the pairwise factors is larger, in both the monolingual and cross-lingual cases. Removing both factors from our proposed model results in a further decrease in the scores. These differences were found to be more significant in the case when $tgt\_size=100$.

Upon looking at the tag set predictions made by our model, we found instances where our model utilizes variable dependencies to predict correct labels. For instance, for a specific phrase in Portuguese (\textit{um estado}), the baseline model predicted \{\textit{POS: Det, Gender: Masc, Number: Sing}\}$_t$, \{\textit{POS: Noun, Gender: Fem (X), Number: Sing}\}$_{t+1}$, whereas our model was able to get the gender correct because of the transition factors in our model.

\subsection{What is the Model Learning?}
\label{sec:analysis}

\begin{figure}[h]
  \centering
  \includegraphics[width=5cm]{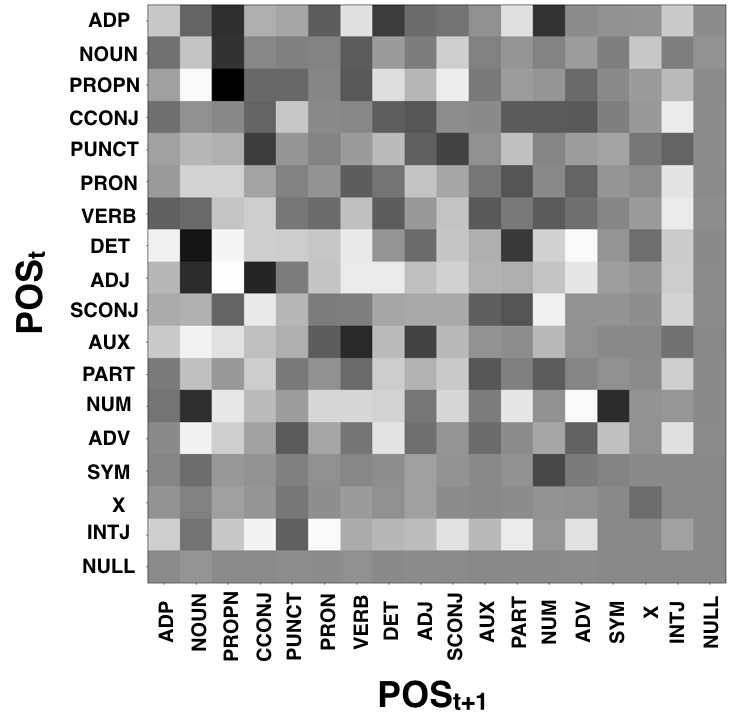}
  \caption{\label{fig:pos} Generic transition weights for \texttt{POS} from the \textsc{Ru/Bg} model}
\end{figure} 

\begin{figure}[h]
  \centering
  \includegraphics[width=5cm]{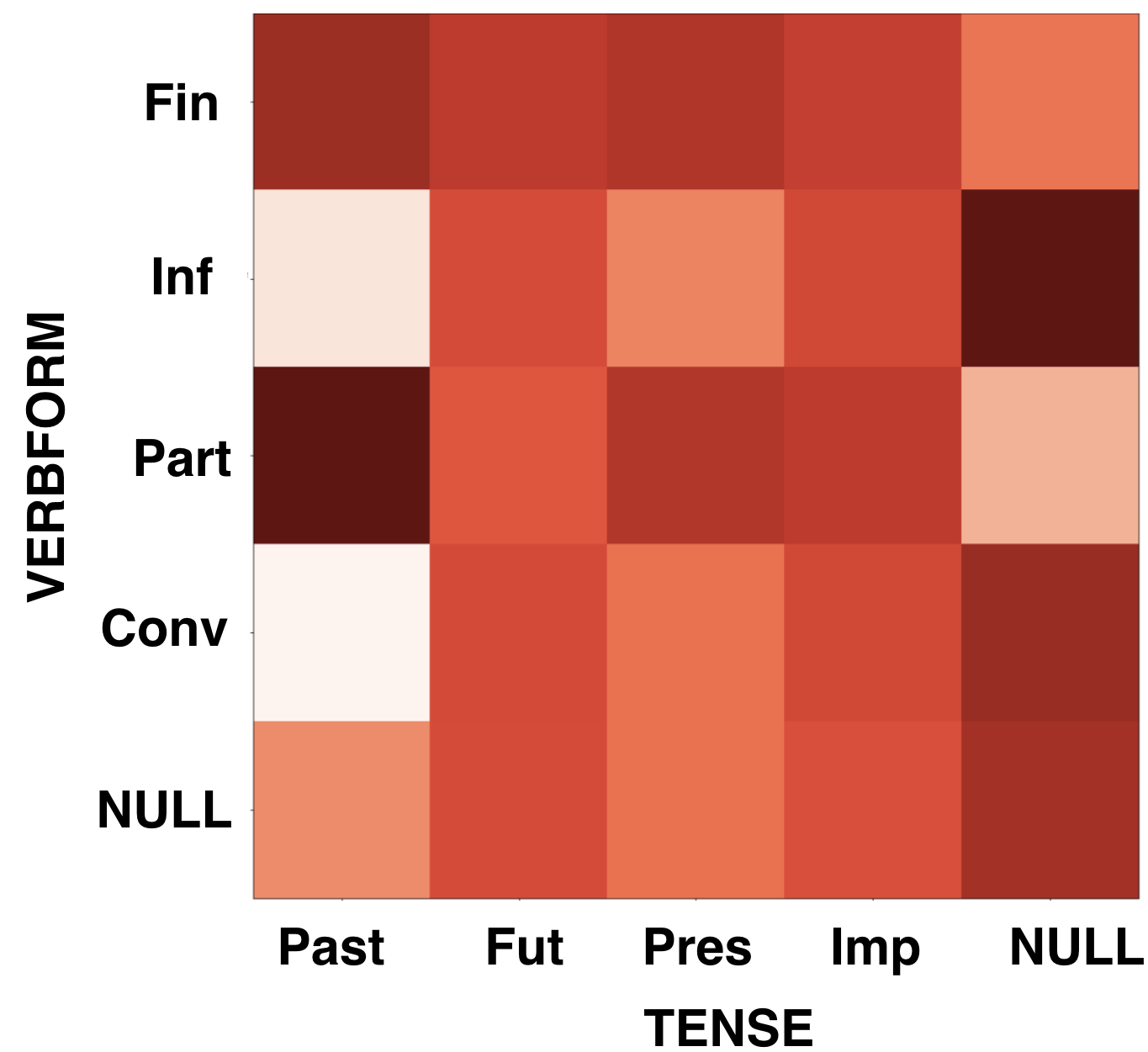}
  \caption{\label{fig:verbform_tense} Generic pairwise weights between \texttt{Verbform} and \texttt{Tense} from the \textsc{Ru/Bg} model}
\end{figure}

One of the major advantages of our model is the ability to interpret what the model has learned by looking at the trained parameter weights. We investigated both language-generic and language-specific patterns learned by our parameters:
\\
\begin{itemize}
\item{\textbf{Language-Generic}: We found evidence for several syntactic properties learned by the model parameters. For instance, in Figure~\ref{fig:pos}, we visualize the generic ($\lambda_{\textnormal{T, gen}}$) transition weights of the POS tags in \textsc{Ru/Bg}. Several universal trends such as determiners and adjectives followed by nouns can be seen. In Figure~\ref{fig:verbform_tense}, we also observed that infinitive has a strong correlation for \texttt{NULL} tense, which follows the universal phenomena that infinitives don't have tense.}

\begin{figure}[h]
  \centering
  \includegraphics[width=5cm]{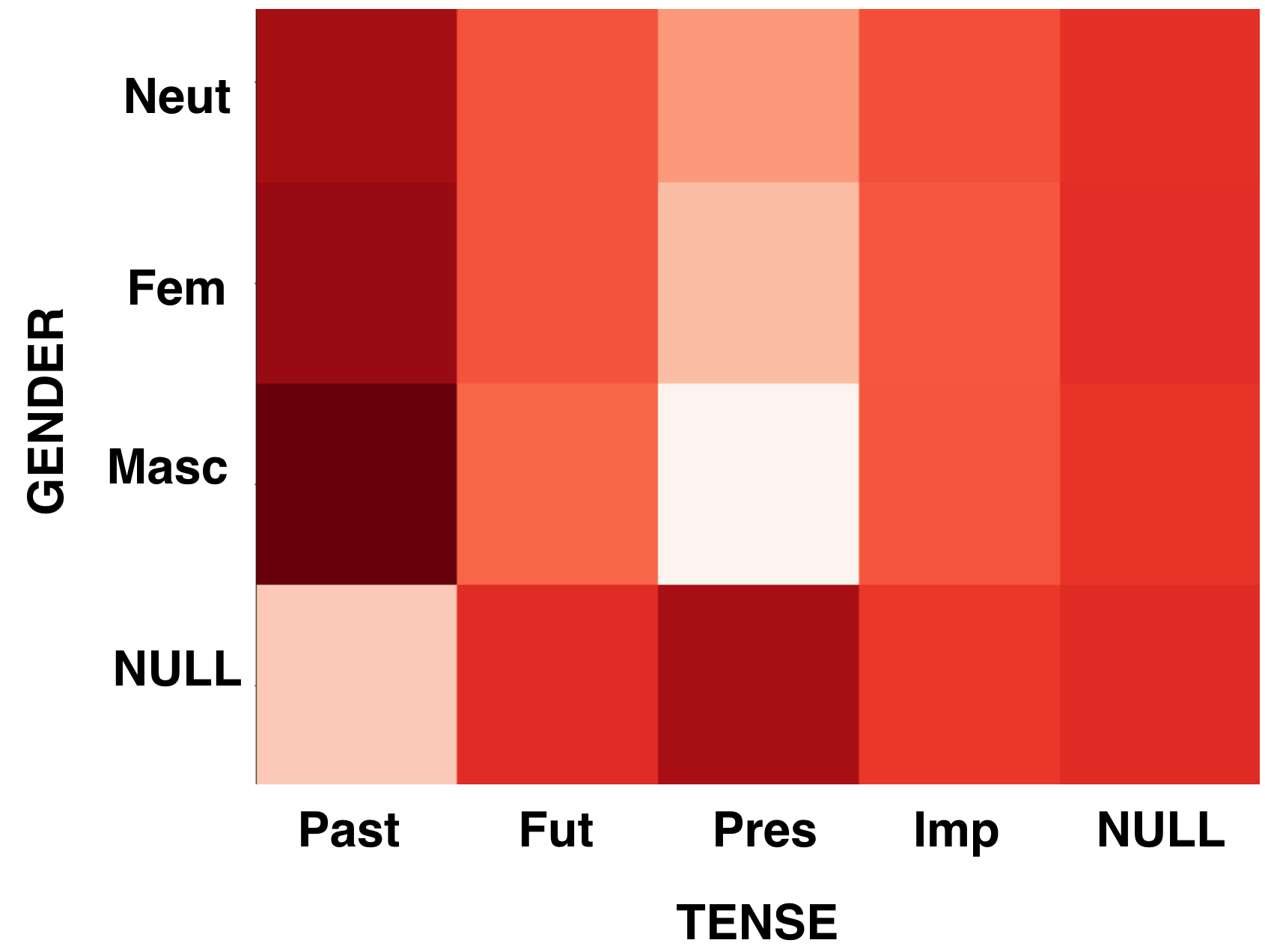}
  \caption{\label{fig:ru_gender_tense} Language-specific pairwise weights for \textsc{Ru} between \texttt{Gender} and \texttt{Tense} from the \textsc{Ru/Bg} model}
\end{figure}

\item{\textbf{Language Specific Trends}: We visualized the learnt language-specific weights and looked for evidence of patterns corresponding to linguistic phenomenas observed in a language of interest. For instance, in Russian, verbs are gender-specific in past tense but not in other tenses. To analyze this, we plotted pairwise weights for Gender/Tense in Figure~\ref{fig:ru_gender_tense} and verified strong correlations between the past tense and all gender labels.}
\end{itemize}

\section{Related Work}
\label{related_work}


There exist several variations of the task of prediction of morphological information from annotated data: paradigm completion \cite{durrett2013supervised, cotterellparadigm2017}, morphological reinflection \cite{cotterell2017sigmorphon}, segmentation \cite{creutz2005morfessor, cotterell2016segmentation} and tagging. Work on morphological tagging has broadly focused on structured prediction models such as CRFs, and neural network models. 
Amongst structured prediction approaches, \citet{lee2011discriminative} proposed a factor-graph based model that performed joint morphological tagging and parsing. \citet{muller2013efficient, muller2015robust} proposed the use of a higher-order CRF that is approximated using coarse-to-fine decoding. \cite{muller2015joint} proposed joint lemmatization and tagging using this framework. \cite{hajivc2000morphological} was the first work that performed experiments on multilingual morphological tagging. They proposed an exponential model and the use of a morphological dictionary. \citet{buys-botha:2016:P16-1, kirov2017rich} proposed a model that used tag projection of type and token constraints from a resource-rich language to a low-resource language for tagging.

Most recent work has focused on character-based neural models \cite{heigold-neumann-vangenabith:2017:EACLlong}, that can handle rare words and are hence more useful to model morphology than word-based models. These models first obtain a character-level representation of a token from a biLSTM or CNN, which is provided to a word-level biLSTM tagger. \citet{heigold-neumann-vangenabith:2017:EACLlong, heigold2016neural} compared several neural architectures to obtain these character-based representations and found the effect of the neural network architecture to be minimal given the networks are carefully tuned. Cross-lingual transfer learning has previously boosted performance on tasks such as translation ~\cite{johnson2016google} and POS tagging~\cite{snyder2008unsupervised,plank2016pos}. \citet{cotterell2017crossling} proposed a cross-lingual character-level neural morphological tagger. They experimented with different strategies to facilitate cross-lingual training: a language ID for each token, a language-specific softmax and a joint language identification and tagging model. We have used this work as a baseline model for comparing with our proposed method.

In contrast to earlier work on morphological tagging, we use a hybrid of neural and graphical model approaches. This combination has several advantages: we can make use of expressive feature representations from neural models while ensuring that our model is interpretable. Our work is similar in spirit to \citet{huang2015bidirectional} and \citet{ma2016end2end}, who proposed models that use a CRF with features from neural models. For our graphical model component, we used a factorial CRF~\cite{sutton2007dynamic}, which is a generalization of a linear chain CRF with additional pairwise factors between cotemporal variables.






\section{Conclusion and Future Work}
\label{sec:conclusion}
In this work, we proposed a novel framework for sequence tagging that combines neural networks and graphical models, and showed its effectiveness on the task of morphological tagging. We believe this framework can be extended to other sequence labeling tasks in NLP such as semantic role labeling. Due to the robustness of the model across languages, we believe it can also be scaled to perform morphological tagging for multiple languages together.

\section*{Acknowledgments}
The authors would like to thank David Mortensen, Soumya Wadhwa and Maria Ryskina for useful comments about this work. We would also like to thank the reviewers who gave valuable feedback to improve the paper. This project was supported in part by an Amazon Academic Research Award and Google Faculty Award.
\bibliography{acl2018}
\bibliographystyle{acl_natbib}

\appendix


\end{document}